\documentclass[10pt,twocolumn,letterpaper]{article}

\usepackage[pagenumbers]{cvpr} % To force page numbers, e.g. for an arXiv version

\usepackage{placeins}
\usepackage{fix-cm}
\usepackage{float}

\definecolor{cvprblue}{rgb}{0.21,0.49,0.74}
\usepackage[pagebackref,breaklinks,colorlinks,allcolors=cvprblue]{hyperref}
\usepackage[accsupp]{axessibility}

\def\paperID{} % *** Enter the Paper ID here
\def\confName{CVPR}
\def\confYear{2026}

\title{Perceptual 3D Simulation With Physical World Modeling}

\author{
\vspace{-0.7cm}
\\
Wanhee Lee$^{1}$\thanks{Equal contribution.}\qquad
Klemen Kotar$^{1}$\footnotemark[1]\qquad
Rahul Mysore Venkatesh$^{1}$\footnotemark[1]\\
Jared Watrous$^{1}$\footnotemark[1]\qquad
Daniel L. K. Yamins$^{1}$ \\
[0.5em]
\fontsize{10.5}{12}\selectfont$^{1}$Stanford University
}

\begin{document}
\maketitle

\begin{abstract}

Predicting how a scene will evolve after a desired 3D transformation from images is a central goal in vision, graphics, and robotics. Yet unlike ideal simulators with full access to 3D geometry and dynamics, real world systems must rely on perceptual inputs and local actions that are inherently partial and incomplete. In this work, we present \textbf{P3Sim}, a physical world modeling system that simulates future scene states under both partial observations and incomplete 3D transformation signals. P3Sim is composed of three interacting components: a learned physical world model, a geometric conditioning module, and a persistent scene memory. The world model interprets perception as probabilistic inference over multimodal scene variables, providing predictions of the distributions of any scene variable conditioned on any combination of others. The geometric conditioning module provides a partial 3D transform signal for conditioning the world model at inference time. The persistent scene memory integrates predictions over time, enabling online updates and consistency under uncertainty. By combining learned inference with explicit geometric structure, P3Sim balances data-driven flexibility with built-in inductive bias. This design yields a flexible perceptual simulator that generalizes across diverse 3D transformation tasks, such as novel view synthesis, object manipulation, and dynamic scene prediction, advancing toward general purpose 3D scene understanding and transformation.
\vspace{-0.5cm}
\end{abstract}

\begin{figure}[h]
\vspace{-0.4cm}
    \centering
    \includegraphics[width=\linewidth]{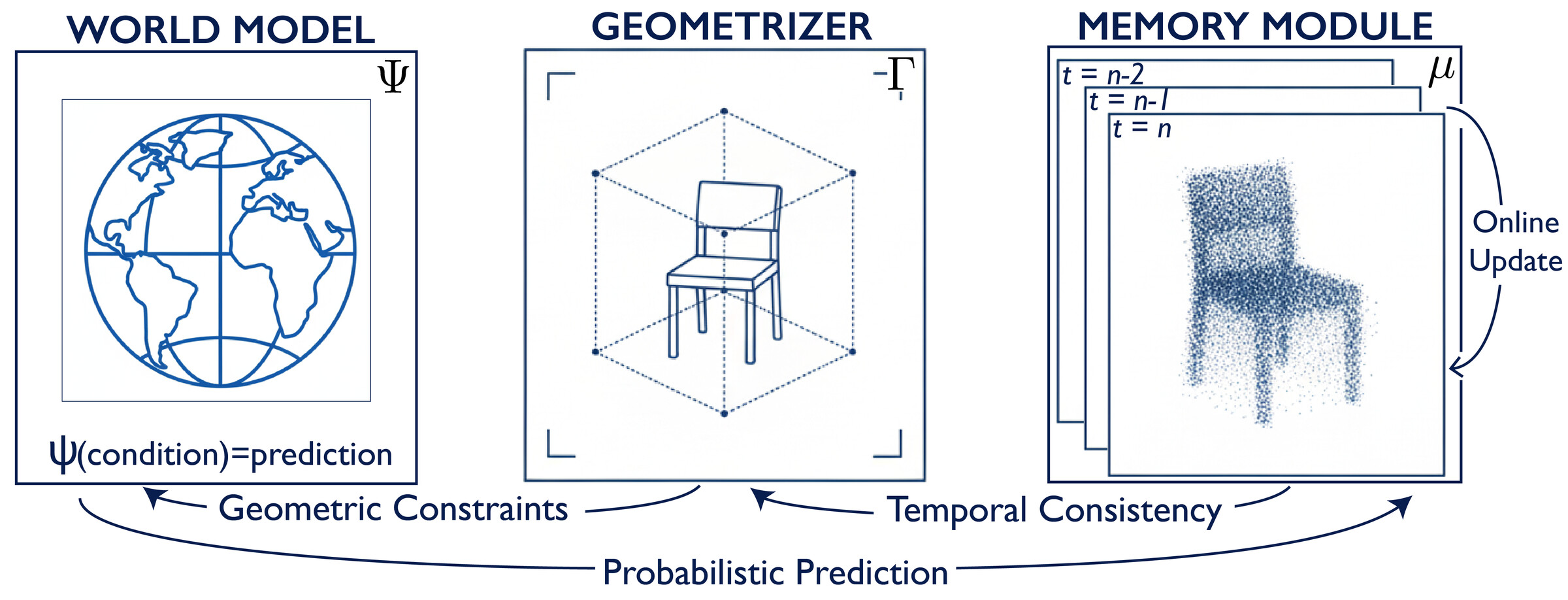}
    \captionsetup{labelfont=bf}
\vspace{-0.6cm}
\caption{
\textbf{\textbf{P3Sim} System Overview.}  
The \textbf{P3Sim} perceptual 3D simulation system consists of three components: a \textit{physical world model} ($\Psi$), a \textit{geometrizer} ($\Gamma$), and a \textit{persistent scene memory} ($\mu$).  
$\Psi$ performs probabilistic inference over multimodal scene variables, RGB, depth, and optical flow, to predict unobserved elements from partial inputs.  
$\Gamma$ derives transformation cues such as partial target depth and motion-consistent flow that describe 3D transforms.  
$\mu$ integrates these predictions over time to maintain a coherent and temporally consistent 3D representation.
    \label{fig:overview}
    }
    \vspace{-0.6cm}
\end{figure}

\vspace{-0.1cm}
\section{Introduction}

Understanding and predicting how a 3D scene evolves under physical or camera transformations is a central goal in vision, graphics, and robotics. In simulated environments such as game engines, this process is straightforward because the system has full access to scene geometry, materials, and physical dynamics. A comparable capability in the real world would enable powerful applications such as controllable video generation, interactive scene manipulation, and embodied reasoning, but achieving this from raw visual inputs remains a major open challenge.

Real-world perception is inherently partial and incomplete, making 3D simulation from images fundamentally difficult. Large parts of the environment are occluded or unobserved, so the full 3D structure cannot be directly computed after an object or camera moves. Moreover, local actions provide only limited information about their global consequences. Lifting one corner of a cloth does not reveal how the rest of it deforms, and pushing an object does not specify whether it will collide or interact with others nearby.
These difficulties highlight three core challenges.
First, reasoning under uncertainty is unavoidable because both the geometry and the transformations are only partially known.
Second, a model must balance built-in geometric structure with learned knowledge from data: some priors, such as projective geometry or motion constraints, are best hard-coded, while others must emerge from experience.
Third, perception and prediction must operate online, continually updating as new observations arrive, which requires an effective mechanism for persistent memory and belief refinement over time.
Our framework explicitly addresses all three challenges through the integration of probabilistic inference, geometric conditioning, and temporal memory.

To address this challenge, we build the Perceptual Physical Simulation system (\textbf{P3Sim}), a unified framework composed of three key components: a data-driven physical world model, a geometric conditioning module, and a persistent scene memory.
The physical world model forms the core of P3Sim, interpreting perception and simulation as probabilistic inference over a graphical model of multimodal scene variables.
Each scene is represented through local scene elements such as RGB patches, depth patches, and optical flow patches, which together describe the visible structure and 3D transforms of the world.
The model can reason about any subset of these variables as conditioning and predict the remaining ones, enabling flexible inference under partial or uncertain observations.
The geometric conditioning module, named geometrizer, provides physically grounded transformation cues by deriving partial depth information in the target frame that represents both scene motion and camera-induced geometry changes.
Finally, the persistent scene memory integrates these predictions over time, maintaining a consistent 3D representation across frames.
Together, these three components, probabilistic inference, geometric conditioning, and temporal memory, enable P3Sim to simulate physically coherent scene evolution from incomplete visual data.

With this framework, we demonstrate a series of 3D scene transformation tasks that illustrate how P3Sim unifies reasoning under both partial geometry and incomplete transformation information.
The system encompasses a broad spectrum of scenarios: it performs 3D scene transformation from incomplete geometry, including novel view synthesis, object manipulation, and joint tasks that combine both viewpoint and object motion.
It also extends to transformations where motion information is incomplete, such as deformable object manipulation, where only part of the object is moved, and multi-object interactions, where moving one object may lead to unknown collisions or secondary motions of nearby objects, as well as multi-agent settings involving humans or vehicles whose actions should be predicted.
Finally, P3Sim leverages its generative capability to complete 3D geometry by integrating predicted states over time, enabling the reconstruction of consistent 3D structure.
Together, these results show that P3Sim effectively addresses the key challenges of uncertainty, inductive bias, and memory, demonstrating a unified approach to physical reasoning and 3D scene transformation from incomplete perceptual data.

\begin{figure*}[!t]
    \centering
    \includegraphics[width=0.98\linewidth]{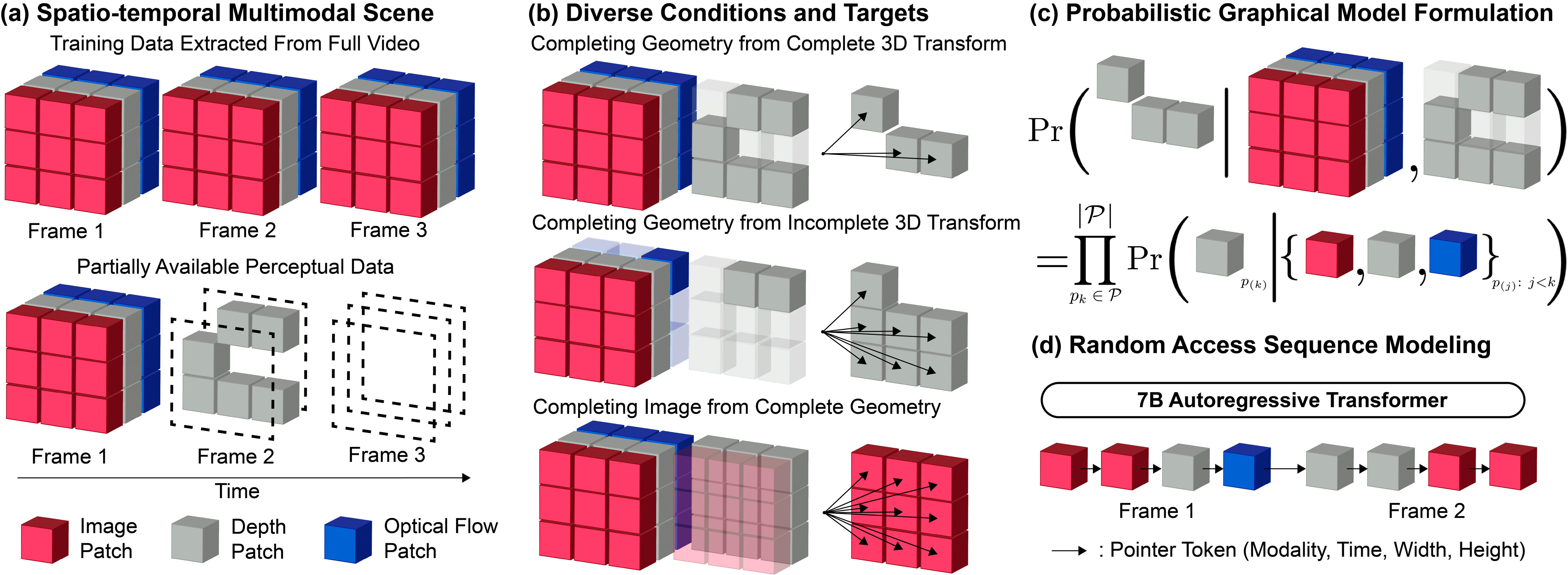}
    \captionsetup{labelfont=bf}
\caption{
\textbf{Physical World Modeling with Multimodal Probabilistic Prediction.}  
Our model simulates 3D scene evolution from partial observations across RGB, depth, flow, and camera pose.  
(a) Multimodal spatio–temporal inputs define partial structure and motion.  
(b) The model flexibly conditions on available cues (e.g., known flow, partial depth) and predicts the rest.  
(c) A probabilistic graphical model represents multimodal scene variables as local nodes.  
(d) Random-access sequence modeling serializes these nodes as pointer–value pairs, enabling controllable inference across space, time, and modality. }
    \label{fig:architecture}
    \vspace{-0.4cm}
\end{figure*}

\vspace{-0.2cm}
\section{Related Works}

\textbf{Camera driven conditioning.}
Novel View Synthesis (NVS) models aim to predict new views from observed images and a target pose. Diffusion-based models such as Zero-1-to-3, ZeroNVS, MotionCtrl, ViewCrafter, and SEVA~\cite{liu2023zero,sargent2024zeronvs,wang2024motionctrl,yu2024viewcrafter,zhou2025stable} improve over regression-based NVS~\cite{yu2021pixelnerf,charatan2024pixelsplat} by sampling diverse and plausible completions in unobserved regions. These methods rely on diffusion pipelines and often lack robust camera control and flexible 3D transforms. Our model uses an autoregressive formulation that supports more robust and broader 3D conditioning.

\textbf{Local motion driven conditioning.}
Object manipulation focuses on transforming selected regions of a scene while the camera remains fixed. Drag based methods such as \cite{wang2024motionctrl, wu2024draganything, shi2024lightningdrag, yin2023dragnuwa} guide stable diffusion \cite{rombach2022high} by converting user inputs into two dimensional motion cues. These approaches can be extended to simple three dimensional modifications by incorporating depth information into the drag signal \cite{objctrl2.5d}. Other models \cite{pandey2024diffusion, koo2025videohandles} edit the input depth map according to the desired transformation and use a depth conditioned diffusion model to synthesize the updated image. These methods rely on inverting the input image into the latent space of stable diffusion which often fails on real images \cite{mokady2023null}, whereas our model avoids latent inversion and performs local edits through the same autoregressive probabilistic inference used for NVS.

\textbf{Tracking based joint motion.}
Recent work explores motion control by conditioning diffusion models on tracked trajectories. Methods such as \cite{geng2025motion, koroglu2025onlyflow, jin2025flovd} train spatio-temporal control networks on top of large video diffusion backbones \cite{bar2024lumiere} and demonstrate abilities that include object movement, camera motion, drag based editing, and motion transfer. Other approaches \cite{gu2025diffusion, zhang2025world, feng2025i2vcontrol} use 3D point trajectories to provide stronger spatial control during generation, though consistent control in complex real scenes remains limited. Our model, in contrast, operates reliably in real world settings and supports arbitrarily dense 3D control signals, while prior methods typically provide only sparse and limited control.

\textbf{World Modeling.}
World models seek to build a unified generative system that can interpret sensory input, predict future states, and support interaction. Early approaches such as Dreamer~\cite{hafner2019dream} learned latent dynamics in simplified settings, while recent large scale models including Genie~\cite{bruce2024genie} and GAIA-1~\cite{hu2023gaia} expand this idea to multimodal, physically grounded domains. However, their action conditioning is often limited and relies on semantic cues that do not enforce geometric consistency. Following recent probabilistic 3D world modeling approaches~\cite{kotar2025world,lee2026unified}, our work predicts 3D scene changes from partial visual evidence, using depth and optical flow to support physically coherent, controllable 3D simulation in real world settings.

\vspace{-0.2cm}
\section{Method}

\begin{figure*}[!t]
% \vspace{-0.1cm}
    \centering
    \includegraphics[width=0.98\linewidth]{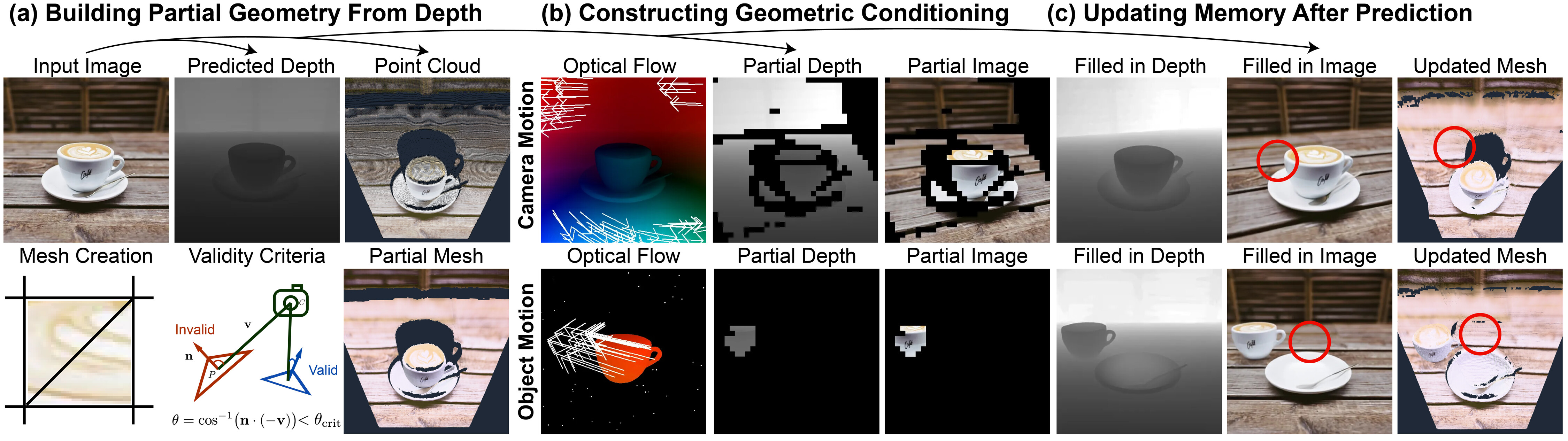}
    \captionsetup{labelfont=bf}
\vspace{-0.2cm}
\caption{
\textbf{Constructing Geometric Conditioning Signals and Persistent Scene Memory.}  
(a) Partial geometry is constructed from reference depth by identifying valid and visible surfaces.  
(b) After applying known 3D transformations, the geometrizer $\Gamma$ reprojects valid regions to the target view to generate partial depth and flow conditioning.  
(c) The persistent scene memory $\mu$ integrates predicted and observed geometry over time, pruning inconsistent or occluded regions to maintain a coherent 3D representation.  
Together, these modules provide consistent geometric evidence for perceptual 3D simulation.
}
    \label{fig:graphics}
    \vspace{-0.6cm}
\end{figure*}

We describe the three main components of our perceptual 3D simulation system: the \textbf{physical world model} $\Psi$, the \textbf{geometric conditioning module} $\Gamma$, and the \textbf{persistent scene memory} $\mu$.
The physical world model (\autoref{fig:architecture}) interprets perception and simulation as probabilistic inference over multimodal scene variables, predicting any unobserved subset given the observed ones.
The geometrizer $\Gamma$ (\autoref{fig:graphics}) provides the world model with transformation cues derived from partial depth and optical flow that represent physically consistent 3D motion and camera-induced geometry changes.
These cues serve as explicit conditioning inputs that guide $\Psi$ in simulating the next scene state under desired transformations.
The persistent scene memory $\mu$ (\autoref{fig:graphics}) integrates predictions from successive frames, storing both observed and inferred variables to maintain a coherent 3D representation over time.
Together, these components enable the system to simulate physically grounded scene evolution from incomplete visual data.

% figure has (a) building partial geometry from depth (b) rendering after 3d transform (c) updating memory after prediction
\vspace{-0.1cm}
\subsection{Physical World Modeling with Multimodal Probabilistic Prediction}

The physical world model ($\Psi$) learns to infer and simulate future 3D scene states from incomplete observations by predicting missing geometric and motion information across modalities within a unified probabilistic framework.

\paragraph{Probabilistic World Model formulation.}
We represent a scene as a set of random variables $\{x_p\}_{p\in\mathcal{P}}$. The set $\mathcal{P}$ enumerates the variable identities, with each \emph{pointer} $p\in\mathcal{P}$ indexing a persistent local scene element, e.g. a spatial location, time index, and modality.  
The set of possible values the variables $x_p$ can take on is denoted by $\mathcal{V}$, representing, e.g. RGB intensities, optical flow displacements, or depth levels.  
At a given time, the currently observed state is denoted by $X$, which is described by a partial function $X:\mathrm{dom}(X)\subseteq\mathcal{P}\rightarrow\mathcal{V}$ mapping each observed pointer $p \in dom(X)$ to its measured value $X(p)$, the observed instance of the variable $x_p$.

The goal of the physical world model $\Psi$ is to infer the conditional distribution of any unobserved variable $x_p$ given the observed state $X$:
\[
\Psi:\,(X,\, p\notin\mathrm{dom}(X)) \mapsto \{\Pr[(p,v)\mid X]\mid v\in\mathcal{V}\}.
\]
This formulation, in effect, defines a \emph{probabilistic graphical model} (PGM) over the scene variables~\cite{koller2009probabilistic}, where inference corresponds to predicting the conditional distributions of missing nodes.  
Sampling from $\Psi$ then produces plausible completions for all unobserved variables that are consistent with the available observations.

\paragraph{Sequence formulation with pointers.}
Directly training the PGM over all scene variables has generally been considered intractable due to the combinatorial number of possible conditioning subsets~\cite{frey2013learning}.  
To make training and inference tractable, we reformulate the model as an autoregressive sequence prediction problem by serializing the data $X$ into an interleaved sequence of pointers \(p_{i}\) and content tokens \(v_{i}\), and train a causal transformer to model the distribution over the content token at an arbitrary query pointer $p$.
\begin{equation*}
\Psi(X \circ p)\;\equiv \Pr\big[v \mid p_{0},v_{0},\ldots,p_{k},v_{k},p\big].
\label{eq:autoregressive_inference}
\end{equation*}
Unlike fixed-order sequential modeling (e.g., raster-order image generation), pointer tokens make the traversal order itself a controllable variable, enabling \emph{random-access} decoding.  
This formulation converts high-dimensional PGM inference into standard GPT-style training over randomized traversal paths, bridging structured probabilistic reasoning with large-scale generative modeling.

\paragraph{Token types and sequence design for 3D control.}
Each scene is represented by multimodal variables \(\{x_p\}_{p\in\mathcal{P}}\), where each pointer \(p\) specifies spatial location, time index, and modality and each variable \(x_p\) takes a discrete value \(v \in \mathcal{V}\) encoding RGB, depth, or flow content.  
A learned encoder maps continuous inputs to compact latent codes (see supplementary materials for details on encoding and quantization).  
Together, \(\mathcal{P}\) and \(\mathcal{V}\) define the token types of our world model, capturing appearance, geometry, and motion.

Each training sample consists of interleaved pointer–value pairs, $(p_0,v_0), (p_1,v_1), \ldots, (p_k,v_k)$.
Randomized pointer order across space lets the model infer missing variables from arbitrary subsets of known ones.  
By choosing modality subsets (RGB, depth, and optical flow) as conditioning tokens, the same formulation unifies geometric reconstruction, novel view synthesis, and motion prediction within a random-access decoding process.

\textit{Training details.}
We train a 7B-parameter autoregressive transformer with next-token prediction cross entropy loss, supervising only the content token (and not the pointer token) sequence elements. The training dataset consisted of
3 million RGB video clips, yielding a total of approximately 1.4 trillion tokens.
Training used a batch size of 512 for 1.5M steps using a Warmup-Stable-Decay schedule. The learning rate peaks at 3e-4 and decays to zero over the final 100K steps. The sequence length is 4096, increased to 8192 for the last 20K steps. We provide full sequence construction procedures and implementation details in the supplementary materials.

\vspace{-0.1cm}
\paragraph{Flexible inference pathways.}
By prompting the model with combinations of optical flow (scene dynamics) and target depth (post-transform geometry), we define several inference pathways.

We organize prompts along two axes:  
(i) \emph{flow density}, indicating how fully dynamics are known, and  
(ii) \emph{depth sparsity}, indicating how much geometry is available after the desired 3D transform.  
This yields two representative categories:

\begin{itemize}
    \item \textbf{Known dynamics} (dense flow) with \textbf{richer target geometry} (less sparse \(D_{\text{tgt}}\)).  
    The motion field is fully specified, and the model renders the outcome consistent with post-transform structure:
    \begin{equation*}
    \vspace{-0.1cm}
    D_1 \sim \Psi(I_0, D_0, F_{0\to1}, D_{1}^{sparse})
    \label{eq:D1_known}
    % \vspace{-0.5cm}
    \end{equation*}
    \begin{equation*}
    I_1 \sim \Psi(I_0, D_0, F_{0\to1}, D_1)
    \label{eq:I1_known}
    \end{equation*}
    Representative tasks include:
    \begin{itemize}
        \item \emph{Novel view synthesis:} dense flow plus sparse or low-resolution target depth to guide parallax and disocclusion.  
        \item \emph{Rigid object manipulation:} dense flow describing rigid motion plus moderately dense target depth constraining shape and contact.
    \end{itemize}

    \item \textbf{Unknown or partial dynamics} (sparse flow) with \textbf{very sparse geometry} (more sparse \(D_{\text{tgt}}\)).  
    Only limited motion and geometry are known, requiring inference of missing dynamics and structure:
    \begin{equation*}
    D_1 \sim \Psi(I_0, F_{0\to1}^{sparse}, D_{1}^{sparse}).
    \end{equation*}
    Representative tasks include:
    \begin{itemize}
        \item \emph{Deformable object manipulation:} sparse flow at contact regions and very sparse depth for global deformation inference.  
        \item \emph{Object interactions:} sparse flow and depth for a subset of objects, requiring collision and response reasoning.  
        \item \emph{Multi-agent actions:} sparse motion cues under uncertainty, requiring probabilistic prediction.
    \end{itemize}
\end{itemize}

Varying flow density and depth sparsity allows a single model to span confident geometry-driven prediction under known dynamics and open-ended inference under partial observations and transform signals.

\subsection{Constructing Geometric Conditioning Signals}
\label{subsec:geometric_conditioning}

Local scene elements, such as RGB, depth, and optical flow, form nodes in a probabilistic model over scene structure and motion.  
Constructing appropriate geometric conditioning signals provides the concrete 3D evidence required for this model to infer the remaining unobserved variables. We therefore introduce a module, $\Gamma$, for producing deterministic geometric constraints given a scene observation and a user- or agent-specified camera and object transformations:
\[
(F_{t-1\rightarrow t},\, D_t^{\text{sparse}}) = \Gamma(\{D_{0:t-1}\},\, K,\, P_t,\, \{\mathcal{T}_t^{(o)}\}),
\]
where $\{D_{0:t-1}\}$ are the previous depth maps, $K$ is the camera intrinsics, $P_t$ is the target camera pose, and $\mathcal{T}_t^{(o)}$ denotes the desired 3D transform for each object $o$.  
Here, $F_{t-1\!\rightarrow\!t}$ is a 2D image-space optical flow computed by applying these transforms to the 3D points of $D_{t-1}$ and reprojecting them into the target frame, while $D_t^{\text{sparse}}$ retains only regions with reliable geometric evidence propagated from prior observations. At inference time, the specified transformations are assumed known by task definition. At training time, depth and flow are estimated and therefore noisy. The outputs of $\Gamma$ serve as conditioning inputs to the $\Psi$, which predicts the remaining uncontrolled aspects of the scene probabilistically.
% The flow $F_{t-1\!\rightarrow\!t}$ is a 2D image-space optical flow computed by applying these transforms to the 3D points of $D_{t-1}$ and reprojecting them into the target frame, while the sparse target depth $D_t^{\text{sparse}}$ retains only regions with reliable geometric evidence propagated from prior observations. The outputs of the geometrizer $\Gamma$, the flow $F_{t-1\rightarrow t}$ and sparse depth $D_{t}^{\text{sparse}}$, serve as conditioning inputs to the world model $\Psi$.
%as shown in Eqs. (\ref{eq:D1_known}) and (\ref{eq:I1_known}).

The scene is first divided into \textit{observed} and \textit{unobserved} regions.  
The observed region corresponds to areas visible from the reference camera, while the unobserved region includes occluded or out-of-view parts.  
The surface geometry derived from $\text{depth}_0$ is further classified as \textit{valid} or \textit{invalid} based on local surface orientation.  
A surface is considered valid if the angle between its surface normal $n$ and viewing direction $v$ satisfies $
\theta = \cos^{-1}(n \cdot (-v)) < \theta_{\text{th}}.
$ Surfaces exceeding this threshold are treated as invalid, as they are likely to be unreliable.

\paragraph{Static scenes.}  
For static settings such as novel view synthesis, the camera moves while the scene remains fixed.  
Partial depth\textsubscript{1} is obtained by projecting valid surfaces from the reference view to the target and retaining only rays that hit visible regions before entering occlusions.  
Flow conditioning is computed by transforming 3D points from $\text{depth}_0$ to the target camera and measuring their 2D displacement.

\paragraph{Dynamic scenes with known motion.}  
When both camera and object motions are known, surfaces from $(\text{rgb}_0, \text{depth}_0)$ are segmented and displaced by their 3D transformations.  
Paired undersurface meshes move with each object to preserve occlusion consistency.  
Rendering this composite geometry defines valid partial depth\textsubscript{1}. Flow is obtained by reprojecting the transformed 3D points into the target frame.

\paragraph{Dynamic scenes with partial motion.}  
If only part of the motion is known, flow conditioning is computed only from known motions and other parts are masked. Also, regions with unknown dynamics are invalid and excluded from partial depth\textsubscript{1}. We provide further details in the supplementary materials.

\subsection{Persistent Scene Memory Module}

The persistent scene memory $\mu_t$ is the integrated 3D scene estimate at time $t$. It stores surface elements, unobserved volumes, camera information, and 3D motion fields for each frame in a global coordinate system and preserves only structures that remain consistent with all observed frames. Queries of the form $\mu_t(t')$, where $t'\in[0,t]$, return the scene geometry at frame $t'$ using the accumulated memory from all frames in $[0,t]$.

At each step, we reconstruct the per-frame geometry
\[
\mathcal{G}_t = \text{BackProject}(D_t, T_t),
\]
where $D_t$ is the depth map of the frame, $T_t$ is the camera pose and intrinsics, and $\mathcal{G}_t$ contains both the visible surface geometry and the unobserved volume implied by occlusions and limited camera coverage.

To handle non-rigid scene transformations such as object manipulation, we denote a 3D motion field input $\mathcal{M}_t$ which transforms the geometry data $\mathcal{G}_t$ into the memory coordinate system. In the case of sparse (\textit{e.g.} single-object) motion conditioning, we mask $\mathcal{M}_t$ as described in \ref{subsec:geometric_conditioning}.

The memory is updated by
\[
\mu_t = \text{Update}_{\mu}\Big(\mu_{t-1}, \mathcal{G}_t, \mathcal{M}_t, D_t, T_t\Big).
\]

The update applies two consistency checks. First, the previous memory is transformed into the current camera view and any unobserved volume that projects closer than the observed depth is removed. Second, the new geometry is projected back into earlier viewpoints and any unobserved volume described by past observations is removed. The surviving elements from both sets are merged to form $\mu_t$. Further details including projection rules and handling of partial visibility appear in the supplementary materials.

\vspace{-0.2cm}
\section{Demonstrating the Perceptual 3D Simulator}

\begin{figure}[!t]
% \vspace{-0.3cm}
\centering
    \includegraphics[width=0.98\linewidth]{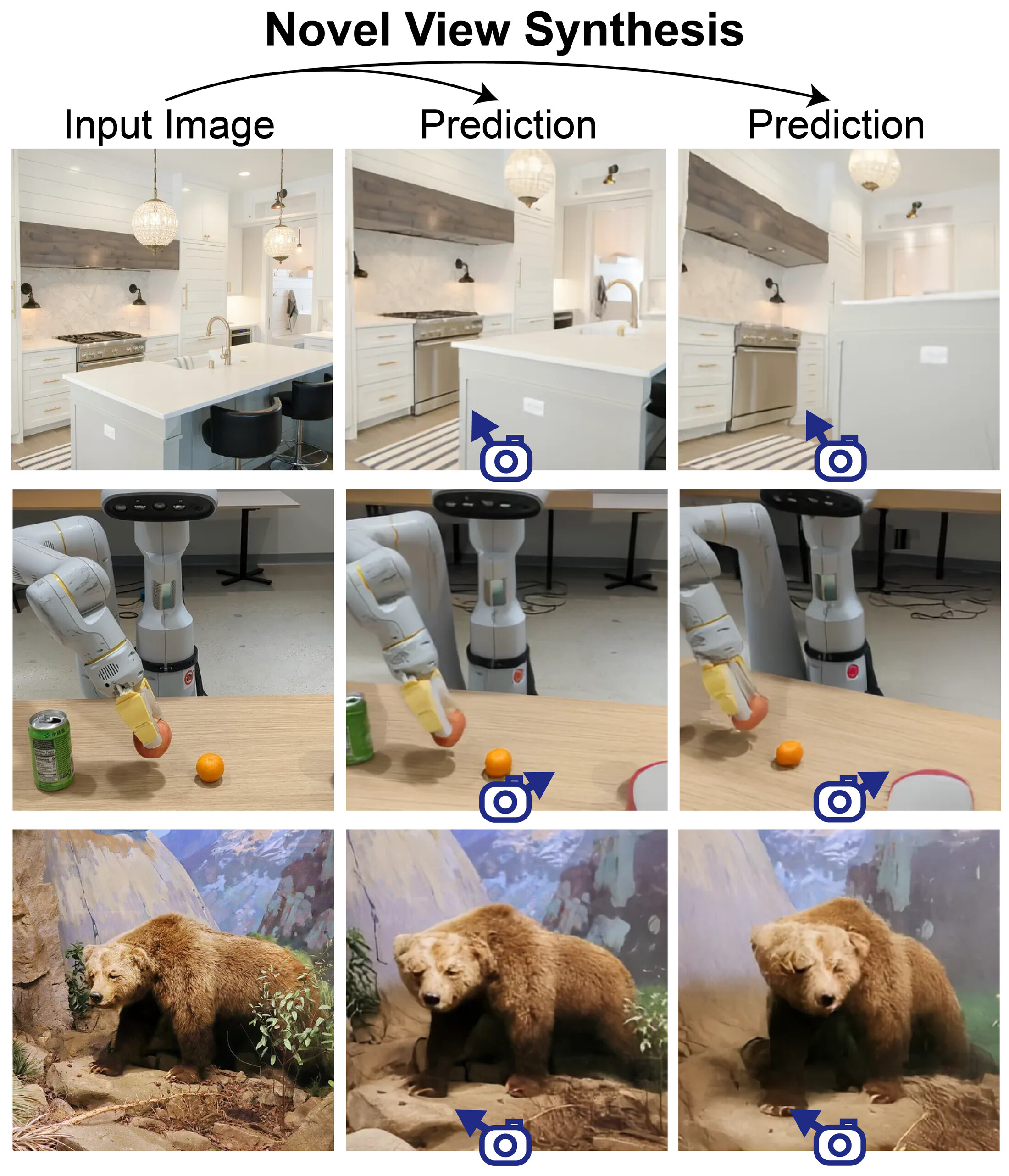}
    \captionsetup{labelfont=bf}
\vspace{-0.4cm}
\caption{
\textbf{Novel View Synthesis.} The model renders unseen viewpoints of static scenes.
This capability supports tasks such as indoor navigation, robot data augmentation, and object-centric reconstruction. \label{fig:nvs}}
\vspace{-0.3cm}
\end{figure}

\noindent
We demonstrate how our model serves as a perceptual 3D simulator that interprets static geometry and dynamic scene evolution directly from visual inputs. First, we show how it infers and manipulates physical state by recovering geometry, synthesizing novel views, and simulating rigid motion (\autoref{fig:nvs}, \autoref{fig:obj}, \autoref{fig:joint}). We then extend this to dynamic phenomena, where the model predicts transformations caused by deformation, collision, and multi-agent interactions (\autoref{fig:deform}, \autoref{fig:collision}, \autoref{fig:other_agent}). Finally, we show that it integrates predictions over time to build a scene-level representation that supports both global and object-centered understanding, including amodal completion of occluded structures (\autoref{fig:extracting_geometry}). These results together illustrate how a unified autoregressive world model can link perception with simulation and support flexible reasoning about physical scenes.

\vspace{-0.1cm}
\subsection{Modeling Physical State: Predicting Geometry}
\begin{figure}[h]
% \vspace{-0.3cm}

    \centering
    \includegraphics[width=0.98\linewidth]{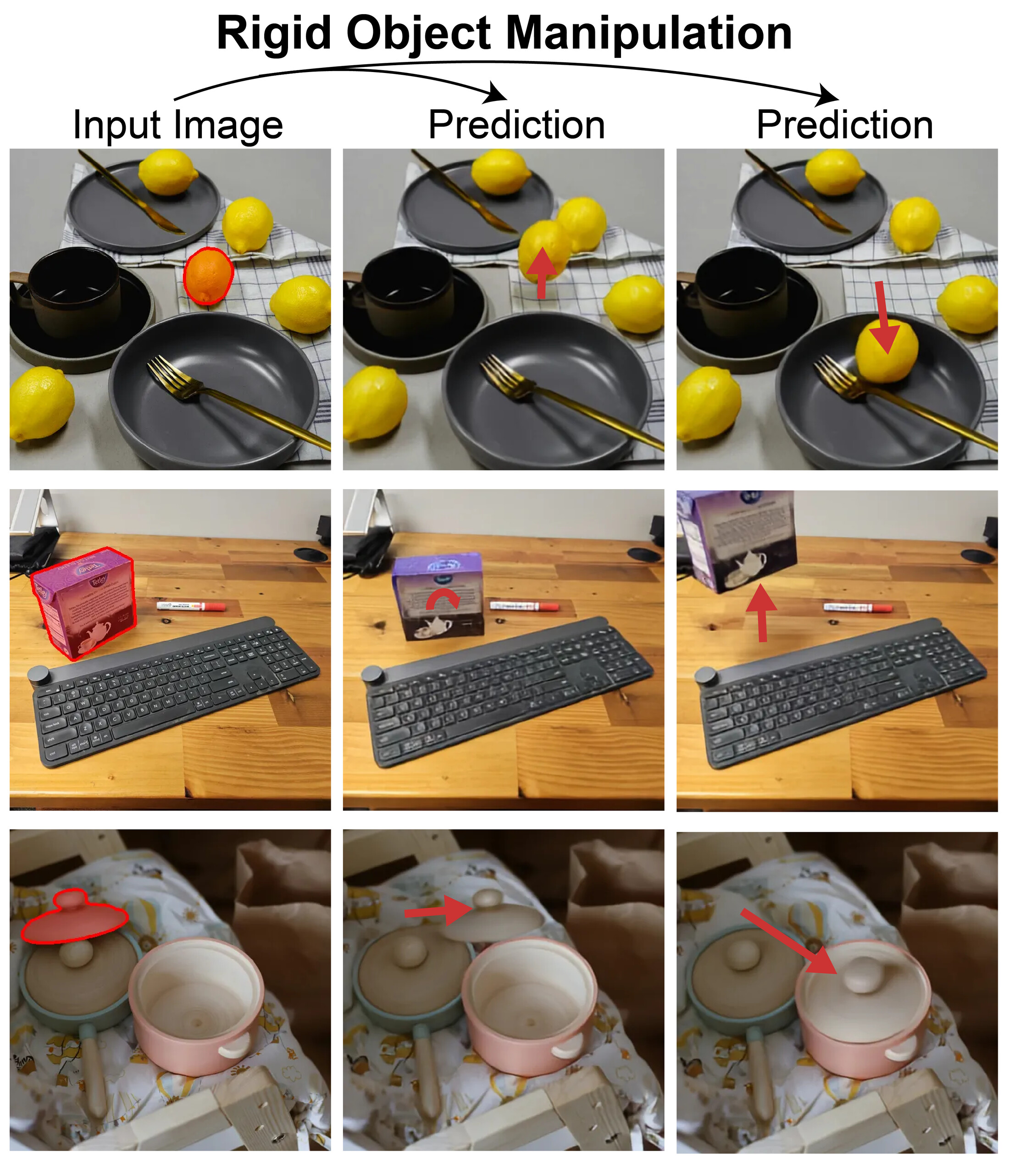}
    \captionsetup{labelfont=bf}
\vspace{-0.4cm}
\caption{
\textbf{Rigid Object Manipulation.} The model simulates scene appearance after a rigid object moves in 3D following a desired transformation.  
This capability can support object manipulation and planning tasks. 
    \label{fig:obj}
    }
    \vspace{-0.3cm}
\end{figure}

\begin{figure}[h]
\vspace{0.1cm}
    \centering
    \includegraphics[width=0.98\linewidth]{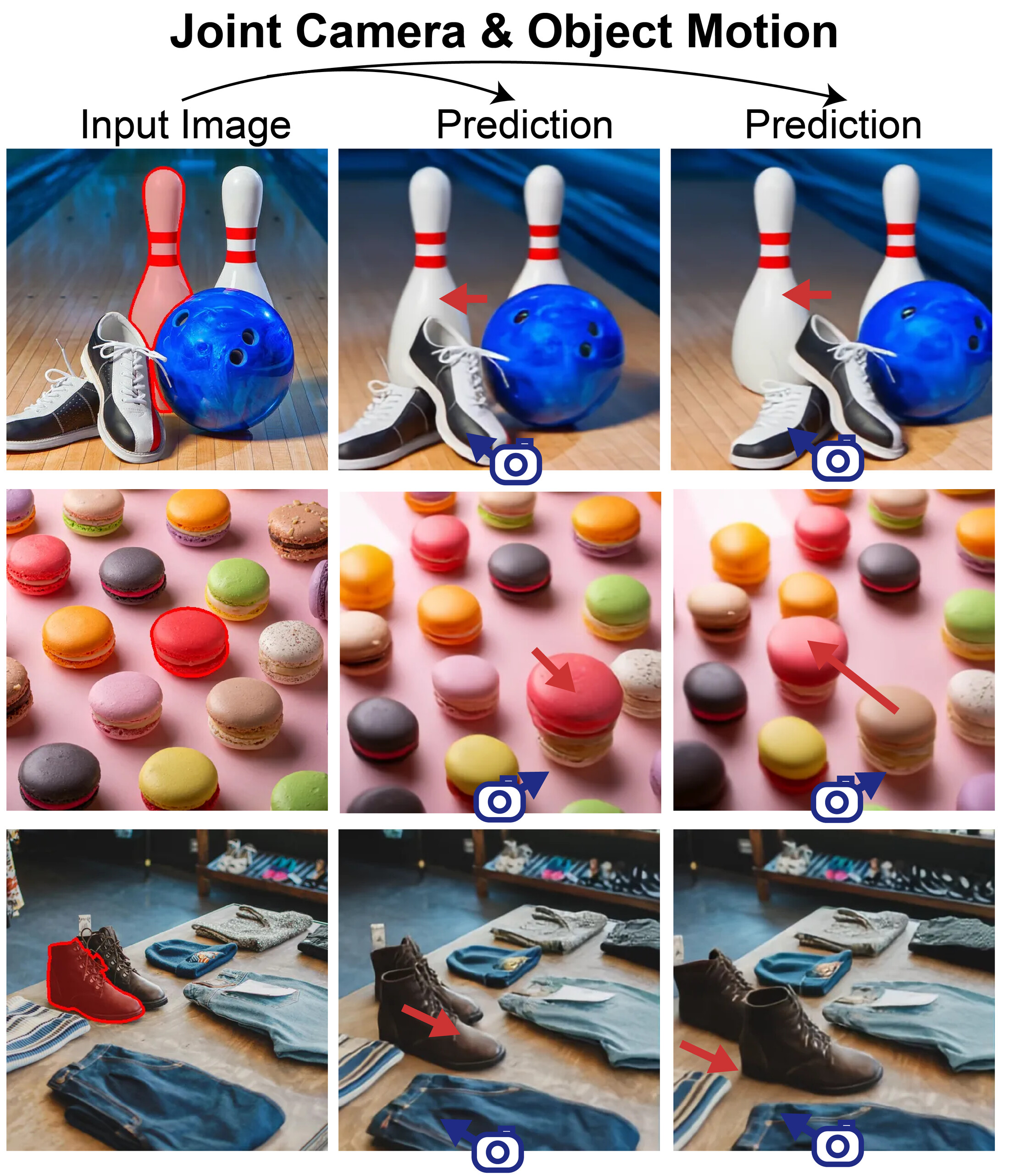}
    \captionsetup{labelfont=bf}
    \vspace{-0.2cm}
\caption{
\textbf{Joint Camera and Object Motion.}  The model predicts scene appearance when both the camera and a rigid object move simultaneously, 
combining global and local 3D transformations.
    \label{fig:joint}
    }
    \vspace{-0.5cm}
\end{figure}

\noindent
We illustrate how the model predicts the scene geometry when the 3D motion of the visible geometry is fully known.

\paragraph{Novel View Synthesis with Simulated Camera Motion}

Novel view synthesis enables navigation and exploration by revealing occluded geometry from new viewpoints (\autoref{fig:nvs}).
In this setting, the scene is static and the camera motion is known, providing dense optical flow and partial target depth as geometric conditioning.

\paragraph{Rigid Object Manipulation}

Rigid object manipulation tests the model’s ability to handle desired object motion (\autoref{fig:obj}).
Known 3D transformations provide dense flow and partial depth signals that define the object’s motion and shape constraints.

\paragraph{Joint Camera and Object Motion}

In scenes where both the camera and multiple objects move, the model integrates global and local displacements (\autoref{fig:joint}).

% \vspace{-0.1cm}
\subsection{Modeling Physical Dynamics: Predicting Transforms}

\begin{figure}[t]
% \vspace{-0.3cm}
    \centering
    \includegraphics[width=0.98\linewidth]{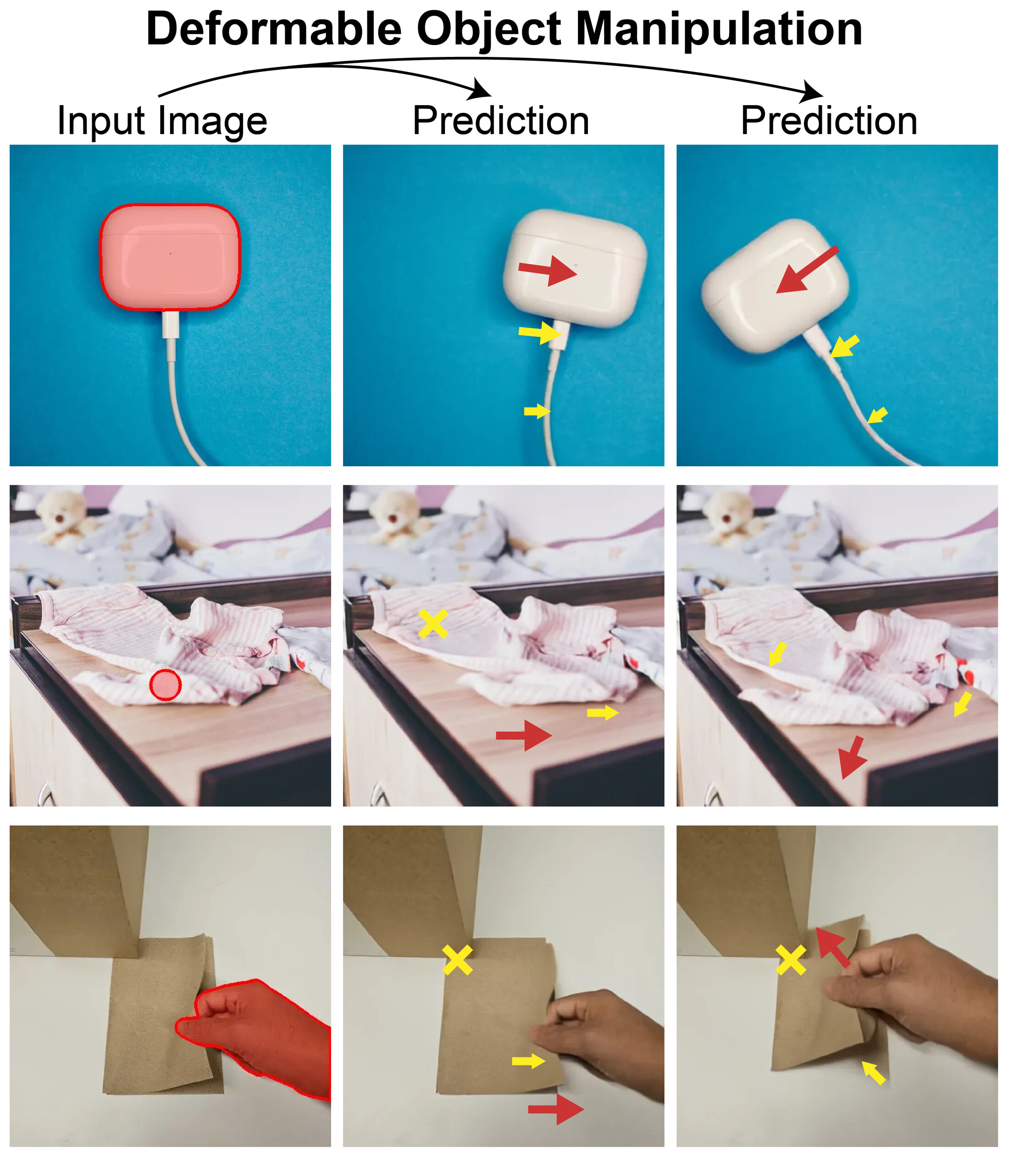}
    \captionsetup{labelfont=bf}
    \vspace{-0.4cm}
\caption{
\textbf{Deformable Object Manipulation.} The model simulates appearance changes when local parts of an object move in 3D, causing other connected regions to deform accordingly.  
Red arrows indicate the input motion, and yellow arrows show the induced motion predicted by the model. 
    \label{fig:deform}
    }
    \vspace{-0.3cm}
\end{figure}

\begin{figure}[h]
% \vspace{-0.3cm}
    \centering
    \includegraphics[width=0.98\linewidth]{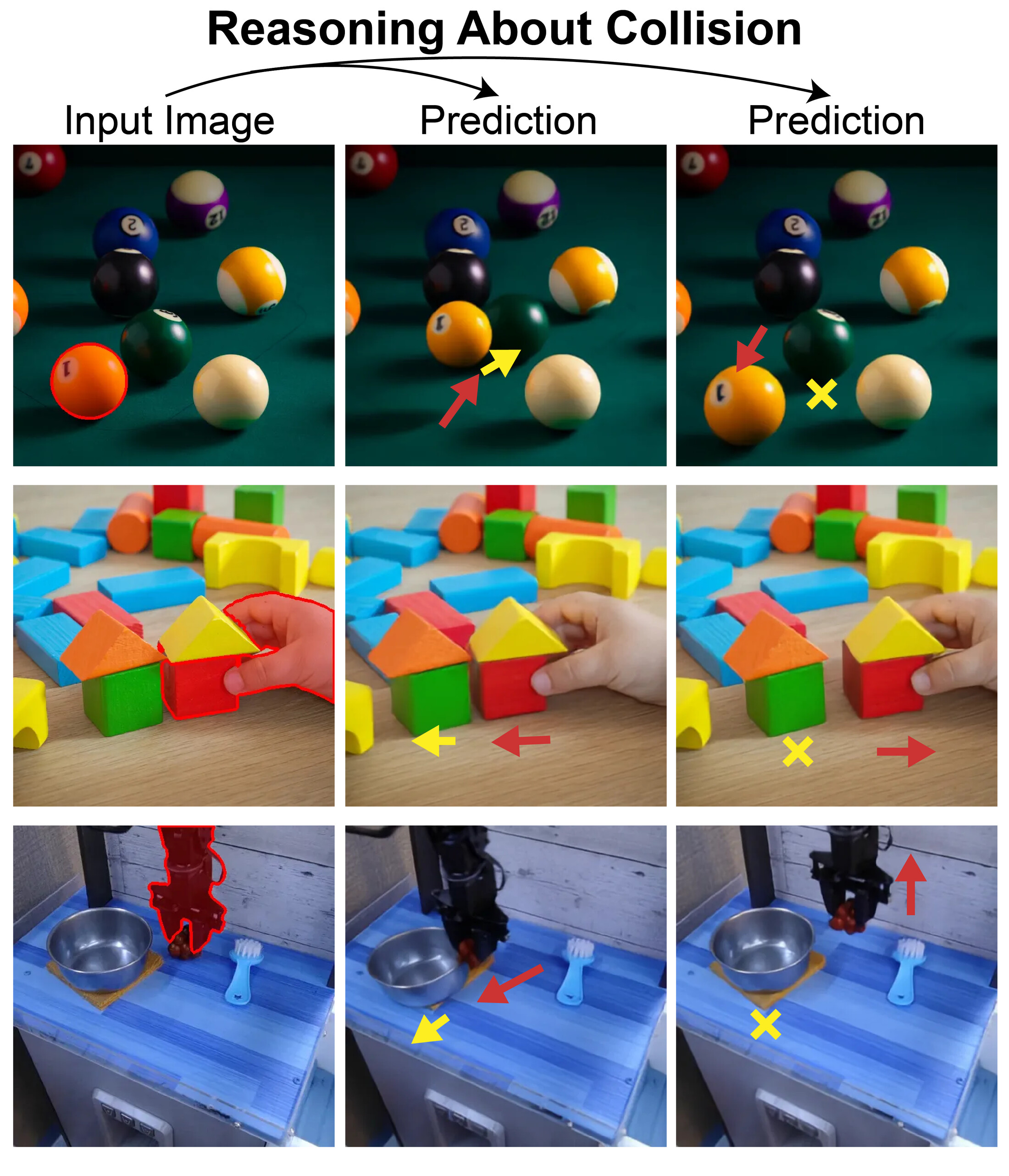}
    \captionsetup{labelfont=bf}
    \vspace{-0.4cm}
\caption{
\textbf{Reasoning about Collision} The model predicts how objects interact when local 3D motion leads to contact or impact in 3D.  
Red arrows indicate the input motion, and yellow arrows show the induced motion predicted by the model. 
    \label{fig:collision}
    }
    \vspace{-0.3cm}
\end{figure}

\begin{figure}[h]
    \vspace{0.1cm}
    \centering
    \includegraphics[width=0.98\linewidth]{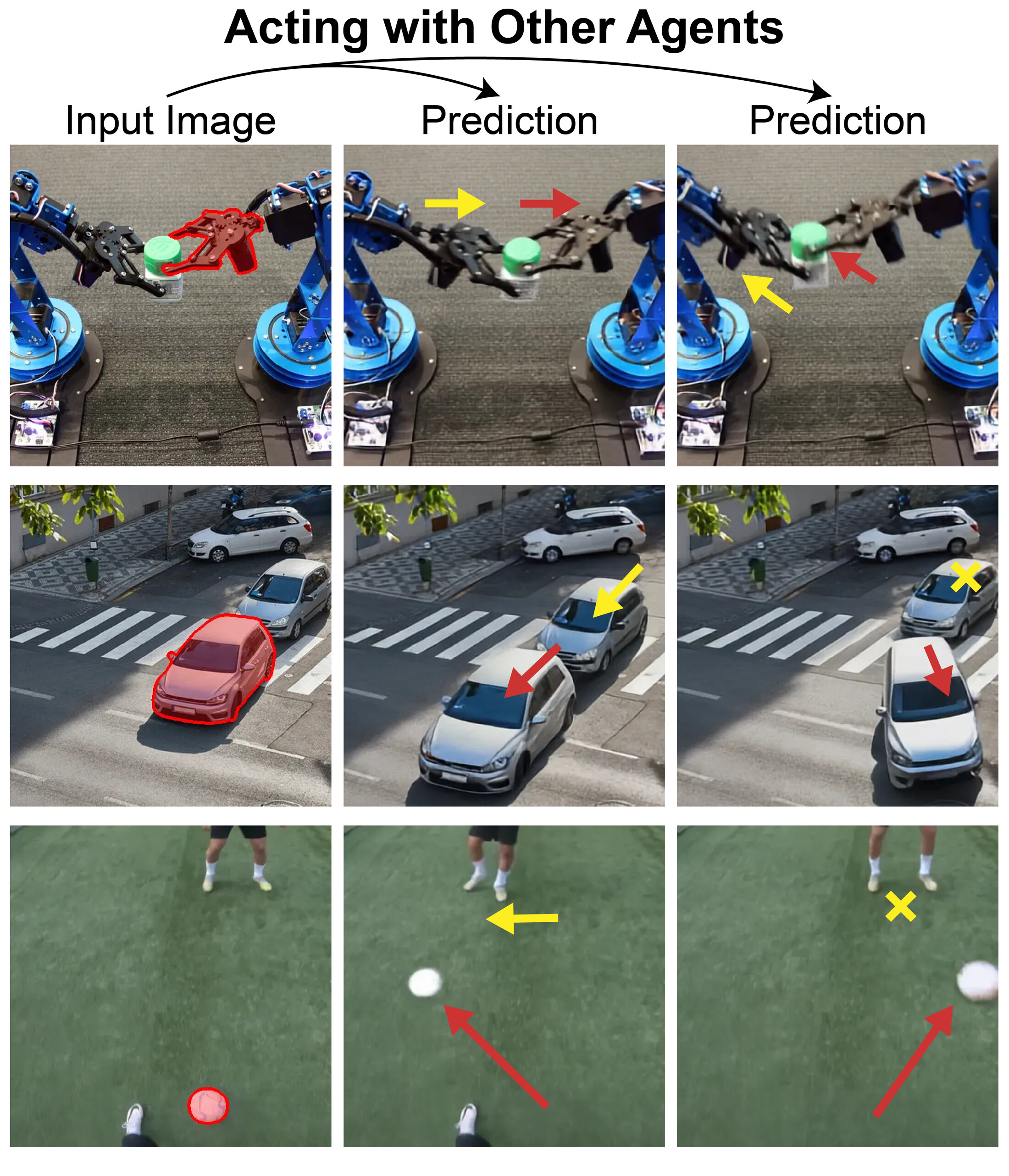}
    \captionsetup{labelfont=bf}
    \vspace{-0.4cm}
\caption{
\textbf{Acting with Other Agents.} The model predicts how an agent’s motion influences and is influenced by another agent in the scene.  
Red arrows indicate the input motion, and yellow arrows show the induced motion predicted by the model.
    \label{fig:other_agent}
    }
    \vspace{-0.9cm}
\end{figure}

\begin{figure*}[!t]
% \vspace{-0.4cm}
    \centering
    \includegraphics[width=0.98\linewidth]{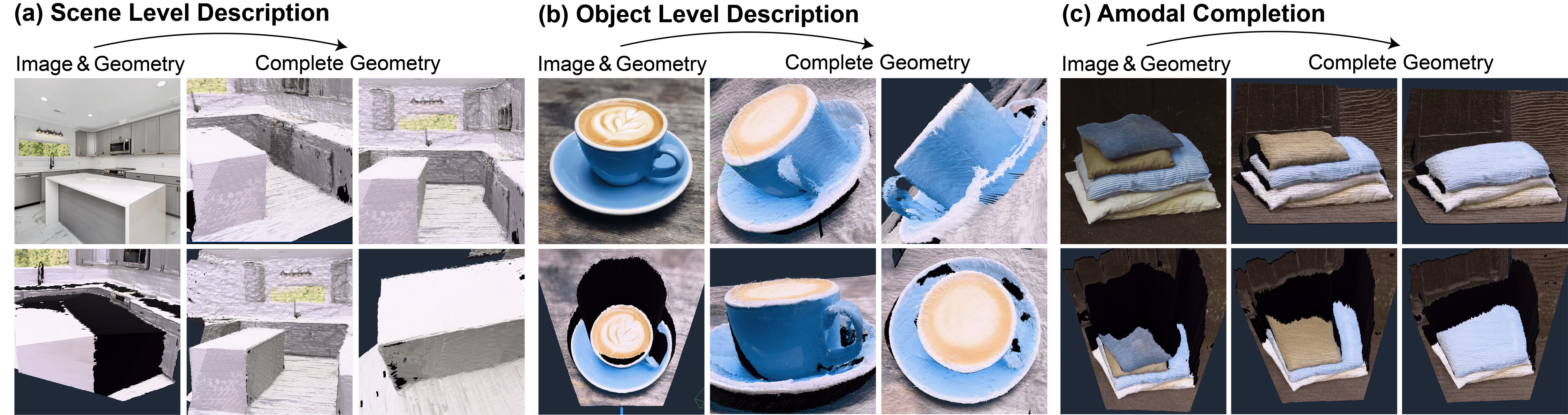}
    \vspace{-0.2cm}
    \captionsetup{labelfont=bf}
    
\caption{
\textbf{Extracting a Complete Scene Description.}  
(a) \emph{Scene-level description:} the model aggregates per-frame geometry into a globally consistent 3D map aligned across views.  
(b) \emph{Object-level description:} object-centric novel view synthesis enables geometrically consistent object shape completion.  
(c) \emph{Amodal completion:} the model reconstructs occluded or unseen structures, yielding a more complete 3D scene representation.
}
    \label{fig:extracting_geometry}
    \vspace{-0.6cm}
\end{figure*}

\noindent
We next explore settings where scene dynamics are only partially known and the model must infer how other objects and materials move.
Here, optical flow conditioning is sparse, and target geometry is highly incomplete, requiring the model to predict missing physical transformations.

\vspace{-0.2cm}
\paragraph{Deformable Object Manipulation}
When only local motion cues at contact regions are known, the model infers global deformation and recovers plausible object shapes after manipulation (\autoref{fig:deform}).

\vspace{-0.2cm}
\paragraph{Reasoning About Collisions}

In scenes with interacting rigid bodies, the model receives partial flow for a subset of objects and must infer collision outcomes for the rest.
These examples suggest that the model can predict plausible physical interactions between objects (\autoref{fig:collision}).

\vspace{-0.2cm}
\paragraph{Acting with Other Agents}

When multiple agents move with uncertain intent, motion cues become ambiguous.
The model can still predict a distribution over possible futures, showing its capacity to represent multi-agent uncertainty under incomplete control (\autoref{fig:other_agent}).

\subsection{Extracting a Complete Scene Description}

\noindent
Finally, we show how the model integrates predictions over time to build a consistent and complete 3D understanding of the scene.
This process uses the persistent scene memory mechanism to merge partial observations into a unified geometric description (\autoref{fig:extracting_geometry}).

\vspace{-0.2cm}
\paragraph{Scene-level Description}
Each frame contributes local geometry reconstructed from its depth and camera pose.
By transforming all frames into a shared coordinate system and pruning geometry inconsistent with current depth observations, the model recovers a coherent scene-level map that maintains visibility consistency across time.

\vspace{-0.2cm}
\paragraph{Object-level Description}
We obtain the object shape through object-centric novel view synthesis. Controlled viewpoint changes around the object reveal surfaces that are not visible in the input frame. Moving the object further helps detach it from the surrounding scene, as demonstrated in the amodal completion result.

\vspace{-0.2cm}
\paragraph{Amodal Completion}
The model infers structure that is never fully visible in any single frame and completes unobserved or occluded regions by removing objects to reveal the hidden geometry.

\vspace{-0.1cm}
\section{Quantitative Results}

\begin{table}[h]
    \centering
    \setlength{\tabcolsep}{5pt}
    \begin{tabular}{lccc}
        \toprule
        \textbf{Model} & \textbf{RE10K} & \textbf{LLFF} & \textbf{DTU} \\
        \midrule
        ViewCrafter  & 20.88 & 10.53 & 12.66 \\
        SEVA         & 18.11 & 14.03 & 14.47 \\
        \textbf{P3Sim (Ours)} & \textbf{21.54} & \textbf{15.18} & \textbf{15.50} \\
        \bottomrule
    \end{tabular}
    \vspace{-0.1cm}
    \captionsetup{labelfont=bf}
    \caption{\textbf{SEVA benchmark PSNR results for NVS.}
    Evaluation under the small-viewpoint NVS setting on the Reconfusion split across RE10K, LLFF, and DTU datasets \cite{zhou2025stable}.}
    \label{tab:nvs}
    \vspace{-0.3cm}
\end{table}

\begin{table}[h]
    \centering
    % \hspace*{-0.4cm} % Adjust the value as needed
    % \renewcommand{\arraystretch}{1.2} % Adjust row height for readability
    \setlength{\tabcolsep}{5pt} % Adjust column spacing
    \begin{tabular}{lccc}
        \toprule
        \textbf{Model} &  \textbf{PSNR}  $\uparrow$ & \textbf{LPIPS} $\downarrow$ & \textbf{EA} $\uparrow$ \\
        \midrule
        DragAnything  &	15.13 & 0.443 & 0.517 \\
        Diffusion Handles & 17.82 &	0.344 & 0.619  \\
        LightningDrag & 19.52 & 0.184 & 0.722  \\        
        \textbf{P3Sim (Ours)} & \textbf{23.12} & \textbf{0.121} & \textbf{0.827} \\
        \bottomrule
    \end{tabular}
    \vspace{-0.1cm}
    \captionsetup{labelfont=bf}
    \caption{\textbf{3D object manipulation results on 3DEditBench.}}
    \label{tab:obj}
    \vspace{-0.3cm}
\end{table}

We quantitatively evaluate our model on two core tasks, novel view synthesis (NVS) and 3D object motion, since these represent the fundamental components of 3D scene transformation. 
Other results in this paper are the natural extensions of these capabilities and do not have a standardized benchmark for evaluation.

\paragraph{Novel View Synthesis.}
For NVS, we use the SEVA~\cite{zhou2025stable}
benchmark, which provides a standardized protocol for single-image view synthesis. We compare our model against ViewCrafter~\cite{yu2024viewcrafter} and SEVA~\cite{zhou2025stable}. Our model achieves high performance across datasets (\autoref{tab:nvs}), which shows that it preserves global scene coherence and provides precise and stable camera control.

\vspace{-0.2cm}
\paragraph{3D Object Manipulation.}
For 3D object manipulation, we evaluate using the \textbf{3DEditBench} benchmark~\cite{lee2026unified}. This dataset provides paired images with known 3D object transformations across diverse categories and scenes. We measure reconstruction quality using PSNR and LPIPS and evaluate edit accuracy with the Edit Adherence (EA) metric \cite{pandey2024diffusion}. We compare against DiffusionHandles~\cite{pandey2024diffusion}, LightningDrag~\cite{shi2024lightningdrag}, and DragAnything~\cite{wu2024draganything}. Quantitative results (\autoref{tab:obj}) show that our model achieves consistently higher PSNR and EA scores, indicating both better image quality and more faithful 3D transformations.

% \vspace{-0.1cm}
\section{Conclusion}

We introduced a framework for perceptual 3D simulation that predicts future scene states under partial observations and incomplete 3D transformations. By extending autoregressive sequence modeling with perceptual 3D signals such as depth and optical flow, the model can simulate camera motion, object interactions, and scene dynamics from a single image. This work advances toward general-purpose physical world models that reason about how visual scenes change over time.

\newpage

\section*{Acknowledgements}
This work was supported by the following awards: To D.L.K.Y.: Simons Foundation grant 543061, National Science Foundation CAREER grant 1844724, National Science Foundation Grant NCS-FR 2123963, Office of Naval Research grant S5122, ONR MURI00010802, ONR MURI S5847, and ONR MURI 1141386- 493027. We also thank the Stanford HAI, Stanford Data Sciences and the Marlowe team, and the Google TPU Research Cloud team for computing support. We thank Yining Hong for helpful discussions that improved the paper.

{
    \small
    \bibliographystyle{ieeenat_fullname}
    \bibliography{main}
}

% \documentclass[10pt,twocolumn,letterpaper]{article}

% % \usepackage[review]{cvpr}  

% \usepackage{cvpr}  

% \usepackage{makecell}
% \usepackage{multirow}
% \usepackage{amsmath}
% \usepackage{graphicx}
% \usepackage{hyperref}
% \usepackage{lineno}
% \usepackage{xspace}  
% \usepackage{amssymb}
% \usepackage{booktabs}   
% \usepackage{subcaption}   
% \def\paperID{} % *** Enter the Paper ID here
% \def\confName{CVPR}
% \def\confYear{2026}
% \begin{document}

% --- Begin Supplementary Formatting ---
\clearpage

% Page numbers: A1, A2, ...
\renewcommand{\thepage}{A\arabic{page}}
\providecommand{\theHpage}{}\renewcommand{\theHpage}{A\arabic{page}}

% --- End Supplementary Formatting ---

% \title{Perceptual 3D Simulation With Physical World Modeling}
% \author{} % CVPR requires empty author in supplementary

% --- Begin Supplementary Formatting ---
\clearpage

% Page numbers: A1, A2, ...
\setcounter{page}{1}
% \renewcommand{\thepage}{A\arabic{page}}

% Reset counters
\setcounter{section}{0}
\setcounter{figure}{0}
\setcounter{table}{0}

% Prefix with A.
\renewcommand{\thesection}{A.\arabic{section}}
\renewcommand{\thefigure}{A\arabic{figure}}
\renewcommand{\thetable}{A\arabic{table}}

\renewcommand{\theHsection}{supp.\arabic{section}}
\renewcommand{\theHfigure}{supp.\arabic{figure}}
\renewcommand{\theHtable}{supp.\arabic{table}}

% Supplementary title
\maketitlesupplementary
% --- End Supplementary Formatting ---

\section*{Overview of the Supplementary Material}

This supplementary material provides additional technical details and qualitative results that complement the main paper.
\begin{itemize}
    \item \textbf{Details of the Physical World Model~$\Psi$}~\ref{sec:psi_details} describes the model architecture, quantizer, and training setup.
    \item \textbf{Details of the Geometrizer~$\Gamma$}~\ref{sec:gamma_details} explains how geometric conditioning signals such as flow and partial depth are constructed.
    \item \textbf{Details of the Persistent Memory Module~$\mu$}~\ref{sec:memory_details} presents the mechanisms for incremental scene reasoning and occupancy update.
    \item \textbf{Additional Examples of Extracting a Complete Scene Description}~\ref{sec:more_geometry_examples} illustrates the model’s behavior in extracting, completing, and stabilizing 3D structure across time.
\end{itemize}

\section{Details of the Physical World Model \texorpdfstring{$\Psi$}{Psi}}
\label{sec:psi_details}

This section provides implementation details of the physical world model $\Psi$. We first describe how local discrete tokens are constructed using Hierarchical Local Quantization (HLQ), followed by how multimodal tokens (RGB, depth, and optical flow) are serialized via Local Random Access Sequences (LRAS). We then explain how a transformer is trained to model their joint distribution. Next, we summarize the sequence-construction rules that mix modalities and temporal contexts, and present a probabilistic interpretation that clarifies how LRAS enables tractable conditional inference. We conclude with the inference settings used during quantitative evaluations.

\paragraph{Hierarchical Local Quantization.}
The world model operates on locally quantized tokens produced by a Hierarchical Local Quantizer (HLQ), a convolutional autoencoder whose receptive field is strictly contained within each patch. This design guarantees that no information propagates across patch boundaries during encoding. Each frame is divided into non-overlapping $16\times16$ patches, and each patch is represented by a short sequence of discrete codes. Each patch contains a coarse token that reconstructs a low-resolution preview, followed by refinement tokens that add high-frequency detail. Because information is confined within a patch, the resulting codes behave predictably under interventions such as masking or overwriting. 

This strict patch-level locality plays a central functional role in controllable inference.
Because no information leaks across patch boundaries during quantization, local interventions, masking a patch, overwriting it with a counterfactual value, or re-ordering its decoding, produce localized and predictable effects. This is essential for the model's ability to perform sparse conditioning, patch regeneration, counterfactual edits, and user-guided generation using arbitrary pointer sequences.

\paragraph{Quantizer training.}
We train separate HLQ quantizers for RGB, depth, and optical flow using a patch-local encoder with a multi-codebook FSQ quantization~\cite{mentzer2024finite} layer and a wavelet input transformation. The decoder is convolutional and nonlocal. All models are optimized on GPUs with a batch size of 512 and a learning rate of $10^{-4}$ until convergence.
RGB quantization uses ImageNet and OpenImages with an $\ell_{1}$ reconstruction loss and a DINOv2 perceptual loss, producing four codes per patch. A coarse head is supervised with an $\ell_{1}$ loss of weight $10^{-2}$ on a $32{\times}32$ target.
Depth maps used for training are generated by VideoDepthAnything~\cite{chen2025video}. The depth quantizer follows the same architecture and losses as RGB but produces two codes per patch.
Optical flow fields are computed using DPFlow~\cite{morimitsu2025dpflow}, with UFM~\cite{zhang2025ufm} added for large-displacement and static-NVS cases. The flow quantizer is supervised with $\ell_{2}$ reconstruction losses and an $\ell_{2}$ coarse loss of weight $10^{-2}$, and likewise produces two codes per patch.
The resulting discrete RGB, depth, and flow tokens form the unified vocabulary consumed by $\Psi$ for multimodal autoregressive modeling.

\paragraph{Autoregressive Transformer.}
The physical world model $\Psi$ is a 7B-parameter decoder-only transformer trained to model the joint distribution over RGB, depth, and optical-flow tokens. Each discrete token is mapped to a learned embedding, augmented with a modality embedding and a spatiotemporal coordinate $(x,y,t,c)$ indicating its patch location, frame index, and channel type. These coordinates are projected into a continuous embedding and added to the token representation before entering the transformer. The backbone uses standard residual-attention blocks with RMS normalization.

Rather than using raster-order flattening, $\Psi$ employs a Local Random Access Sequence (LRAS) that randomly permutes patch locations during training and inference. LRAS exposes tokens to diverse spatial contexts, improves robustness under partial observations, and enables spatially local conditioning, which is an essential capability because geometric signals such as flow or partial depth are sparse and may occur at arbitrary locations.

\paragraph{Sequence Model Training.}
We train $\Psi$ with next-token prediction cross-entropy loss, supervising only the content tokens and not the associated pointer tokens. The training corpus contains approximately 3 million video clips, corresponding to roughly 1.4 trillion multimodal tokens. The dataset spans indoor, outdoor, egocentric, and object-centric settings, combining large-scale sources (e.g., ScanNet++ \cite{yeshwanth2023scannet++}, CO3D \cite{reizenstein2021common}, RealEstate10K \cite{zhou2018stereo}, MVImgNet \cite{yu2023mvimgnet}, DL3DV \cite{ling2024dl3dv}, and EgoExo4D \cite{grauman2024ego}) with curated internet videos selected for scene diversity and long-range dynamics. Training is conducted entirely on a TPU v5e–256 pod through the Google TPU Research Cloud (TRC). We use a global batch size of 512 for a total of 1.5M optimization steps. Each training step requires approximately 3.8 seconds. The learning rate follows a Warmup–Stable–Decay schedule, increasing to a peak value of $3 \times 10^{-4}$, remaining constant during the stable phase, and decaying to zero over the final 100K steps. The default sequence length is 4096 tokens, and we increase it to 8192 for the final 20K training steps to improve the model’s ability to generate longer rollouts. All training uses mixed precision, FSDP sharding, and gradient checkpointing.

\paragraph{Sequence Design.}
To support a unified autoregressive model capable of reasoning under heterogeneous perceptual inputs, we design a sequence construction pipeline that mixes RGB, depth, and optical flow tokens across a wide range of temporal configurations. Each training sample is produced by selecting a mode that specifies (i) the set of modalities to include, (ii) the number of frames (two to four), and (iii) the temporal spacing. Given the chosen mode, the dataset dynamically loads the quantized patch-level token blocks of RGB, depth, and optical flow. 

A constrained resource allocator then assigns token budgets to each modality–frame pair, using hard capacity limits to ensure that the combined sequence fits within the global token limit (4096 or 8192 tokens). This allocation determines how many patches to keep for each signal before shuffling. Frame indices are sampled according to the specified gap regime, and each patch is annotated with a rotary-position index that encodes spatial location, temporal position, and channel group~\cite{su2024roformer}. Depth tokens may appear before or after RGB tokens, and flow tokens are placed between the RGB frames they connect.

During target construction, the first frame and the pointer tokens are never supervised. The resulting sequence is a heterogeneous but structurally consistent stream of pointer–value tokens that exposes the model to diverse inference conditions, ranging from fully observed RGB sequences to diverse modality scenarios. This diversity is essential for teaching the model to perform general-purpose probabilistic inference over multimodal 3D scene variables.

\paragraph{Probabilistic graphical model interpretation.}

In our formulation, each pointer token $p \in \mathcal{P}$ identifies a scene variable $x_p$ whose value lies in $\mathcal{V}$. A pointer--value pair $(p_i, v_i)$ therefore corresponds to observing or generating the event $x_{p_i}=v_i$. The full collection $\{x_p\}_{p\in\mathcal{P}}$ defines an implicit probabilistic graphical model over multimodal scene elements, including RGB, depth, and optical flow. A Local Random-Access Sequence (LRAS) specifies an ordering over these variables. As the sequence progresses, previously observed or generated pairs $(p_i, v_i)$ become conditioning evidence for predicting the value at a query pointer $p$. The model computes
\[
\Psi(X \circ p)
\;=\;
\Pr\!\big[\,x_p = v \,\mid\, p_0,v_0,\ldots,p_k,v_k,p\,\big],
\]
or equivalently the distribution over the content token $v$ associated with the query pointer $p$. This provides a tractable approximation to conditional inference over $\{x_p\}_{p\in\mathcal{P}}$, avoiding explicit global inference over exponentially many conditioning subsets. More generally, for any observed subset $S \subset \mathcal{P}$ and query pointer $p \in \mathcal{P}\setminus S$, the model can approximate conditionals of the form
\[
\Pr(x_p \mid x_S)
\]
by providing the observed pointer--value pairs $(s, X(s))$ for $s \in S$ as evidence and decoding the remaining variables. This capability enables a broad family of conditional inference pathways, including the flow--depth prompting regimes described in the main paper for novel-view synthesis, rigid and deformable object manipulation, object interactions, and motion prediction. In general, different subsets of multimodal observations can be supplied as evidence, allowing a single sequence model to adapt its inference behavior within a unified probabilistic architecture.

\paragraph{Inference settings.}
All evaluations use sequential decoding with a temperature of $1.0$, top-$p = 0.9$, and top-$k = 1000$. Unless otherwise specified, single-image depth is predicted using the metric-depth model Depth Anything V2~\cite{yang2024depth}. The metric depth is then converted to disparity and normalized so that the conditioning disparity has a maximum scale of 0.6. 

For novel view synthesis (NVS) evaluation, we use Depth Anything 3 because we empirically found that Depth Anything V2 tended to produce more distorted object shapes on DTU in our setup. We additionally estimate a camera pose translation scale factor per scene using UFM and single-view depth from Depth Anything 3~\cite{lin2025depth}. Specifically, we compute optical flow by warping the predicted depth using the ground-truth camera pose under candidate scales, compare it against UFM flow on covisible regions of the ground-truth video, and optimize the scalar scene scale that minimizes this flow discrepancy. This estimated scale places the ground-truth camera motion in the same metric space as the single-view predicted depth. 

\section{Details of Geometrizer \texorpdfstring{$\Gamma$}{Gamma}}
\label{sec:gamma_details}

In this section, we describe the details of the geometrizer $\Gamma$.
Given a depth history $\{D_{0:t-1}\}$, camera intrinsics $K$, a target camera pose $P_t$, and object-level SE(3) transformations $\{\mathcal{T}_t^{(o)}\}$, the geometrizer constructs geometric conditioning signals
\[
(F_{t-1\rightarrow t},\, D_t^{\mathrm{sparse}}).
\]
Here, $F_{t-1\rightarrow t}$ is the induced optical flow and $D_t^{\mathrm{sparse}}$ is the rendered sparse depth. These signals convert the specified camera motion and per-object transformations into sparse depth and optical-flow fields that constrain the world model during autoregressive decoding.

The object transformations $\{\mathcal{T}_t^{(o)}\}$ are defined with respect to user-provided segmentations. Each object’s segmentation mask identifies the pixels belonging to that object, allowing its motion to be applied independently during geometry rendering.

\paragraph{Surface and undersurface meshes per object.}

We begin by reconstructing the reference-frame geometry
\[
\mathcal{G}_0 = \mathrm{BackProject}(D_0,\, T_0),
\]
where $T_0$ contains the reference camera pose and intrinsics.  
For each pixel \(u = (u_x, u_y)\), let \(X_0(u) \in \mathcal{G}_0\) denote the corresponding 3D point.

We partition this geometry according to the segmentation mask:
\[
\mathcal{G}_0^{(o)}
= \{\,X_0(u) \;|\; \mathrm{seg}_0(u)=o\,\},
\]
for each object label $o$, including background.

\textit{Surface meshes.}
For each object $o$, we construct a surface mesh $\mathcal{S}_0^{(o)}$ by triangulating adjacent pixels whose segmentation label equals $o$. We compute triangle normals $n(u)$ and apply the surface-orientation criterion
\[
\theta(u)
= \cos^{-1}\!\left( n(u)\cdot(-v(u))\right)
< \theta_{\mathrm{th}},
\qquad \theta_{\mathrm{th}} = 80^\circ,
\]
where $v(u)$ is the viewing direction. Only triangles satisfying this condition are retained.

\textit{Undersurface meshes.}
We also associate each object $o$ with an undersurface mesh $\mathcal{U}_0^{(o)}$ that parameterizes the locally unobserved volume behind the visible scene.
To initialize these meshes, we first build a single occupancy volume from the full-frame depth map $D_0$, assigning observed space the value $+1$ and unobserved space the value $-1$.
We carve this volume using an offset depth map $D_0(u) + \varepsilon$ (with $\varepsilon = 3\times10^{-2}$), ensuring that the resulting undersurface remains strictly behind the observed surface. Extracting the $0$-level isosurface via marching cubes then yields a global undersurface mesh $\tilde{\mathcal{U}}_0$ that lies just behind the visible surfaces ${\mathcal{S}_0^{(o)}}$ along each viewing ray.
Because the structure of the unobserved volume is not yet known, every object (including the background) is initialized with a copy of this geometry,
\[
\mathcal{U}_0^{(o)} = \tilde{\mathcal{U}}_0 \quad \forall o.
\]
As the scene evolves and objects move, each $\mathcal{U}_t^{(o)}$ is updated independently based on the object motion and new depth observations, so the undersurface meshes gradually diverge.

Each object (including background) thus starts with paired geometry
\[
\bigl(\mathcal{S}_0^{(o)},\, \mathcal{U}_0^{(o)}\bigr),
\]
where the undersurface meshes share the same initial shape but are subsequently deformed and carved in an object-specific manner.

\paragraph{Composite geometry and object motion.}

At time $t$, we apply the specified object motions:
\[
\mathcal{S}_t^{(o)} = \mathcal{T}_t^{(o)}\, \mathcal{S}_0^{(o)},
\qquad
\mathcal{U}_t^{(o)} = \mathcal{T}_t^{(o)}\, \mathcal{U}_0^{(o)}.
\]

The composite scene geometry is the union
\[
\mathcal{C}_t
= \bigcup_{o}\left(\mathcal{S}_t^{(o)} \cup \mathcal{U}_t^{(o)}\right),
\]
ensuring that each object's surface and undersurface geometry move consistently with its SE(3) transformation.

\paragraph{Rendering sparse target depth.}

Rendering $\mathcal{C}_t$ from the target pose $P_t$ produces a depth value $z(u)$ for rays that hit a surface mesh.  
We define
\[
D_t^{\mathrm{sparse}}(u)
=
\begin{cases}
z(u), & \text{if the first hit is a surface mesh},\\
\text{invalid}, & \text{otherwise}.
\end{cases}
\]

In practice, surface meshes are rendered in white and undersurface meshes in black. Any pixel that renders black is marked invalid.  
A patch is invalid if any of its pixels is invalid.  
We use \texttt{pyrender}, though any renderer is suitable.

\paragraph{Computing optical flow.}

We reconstruct the previous-frame geometry
\[
\mathcal{G}_{t-1} = \mathrm{BackProject}(D_{t-1},\, T_{t-1}),
\]
and let $X_{t-1}(u)\in\mathcal{G}_{t-1}$ be the 3D point associated with pixel $u$.

Its object label is $\mathrm{seg}_{t-1}(u)$, so the corresponding SE(3) motion is $\mathcal{T}_t^{(\mathrm{seg}_{t-1}(u))}$.  
We apply this motion and reproject:
\[
X_t(u) = \mathcal{T}_t^{(\mathrm{seg}_{t-1}(u))}\, X_{t-1}(u),
\]
\[
u_t = \mathrm{Project}\!\left(P_t\, X_t(u),\,K\right).
\]

The optical flow is
\[
F_{t-1\rightarrow t}(u) = u_t - u.
\]

Pixels with unknown motion (e.g., unlabeled segments or masked regions) are excluded from the conditioning.

\paragraph{Motion scenarios.}
We summarize how $\Gamma$ constructs conditioning signals for different motion cases.

\textit{Static scenes (novel view synthesis).}
Only the camera pose changes while all objects remain fixed.
The surface and undersurface meshes remain at their reference-frame positions, and the composite geometry is rendered directly from $P_t$.
Valid partial depth$_1$ corresponds to rays whose first intersection is a surface in the reference geometry, excluding occluded or undersurface hits.
Flow is obtained by transforming reference-frame 3D points $X_0(u)$ into the target camera and measuring their 2D displacement.

\textit{Dynamic scenes with fully known motion.}
When both camera and object-level SE(3) motions $\{\mathcal{T}_t^{(o)}\}$ are provided, each mesh pair $(\mathcal{S}_0^{(o)}, \mathcal{U}_0^{(o)})$ is displaced independently:
\[
\mathcal{S}_t^{(o)}=\mathcal{T}_t^{(o)}\mathcal{S}_0^{(o)}, \qquad
\mathcal{U}_t^{(o)}=\mathcal{T}_t^{(o)}\mathcal{U}_0^{(o)}.
\]
Rendering the resulting composite geometry $\mathcal{C}_t$ from $P_t$ defines partial depth$_1$, and flow is computed by applying the corresponding object motion to each previous-frame point $X_{t-1}(u)$ before reprojecting into the target view.

\textit{Dynamic scenes with partially known motion.}
If only a subset of object motions is specified, partial depth$_1$ is constructed only from the objects with known SE(3) transformations.
These objects have their surface meshes transformed and rendered, and all other surfaces are treated as unknown.
Rays whose first hit belongs to an object with unknown motion are marked invalid and do not contribute to partial depth.
Flow is defined in the same way.
Pixels whose object labels correspond to unknown motion are masked, and the remaining regions follow the procedure used in the fully known-motion setting.

% -----------------------------------------------------
\section{Details of Persistent Memory Module \texorpdfstring{$\mu$}{mu}}
\label{sec:memory_details}
This section provides an implementation of the abstract memory-update rules described in the main paper, specifying how the memory is represented using surface meshes and occupancy-based undersurface meshes. The persistent memory $\mu_t$ maintains, for each object $o$, a surface mesh $\mathcal{S}_t^{(o)}$ and an undersurface mesh $\mathcal{U}_t^{(o)}$ derived from an occupancy volume that tracks unobserved space. All geometry is stored in a globally aligned reference frame and updated using object-level SE(3) transformations.

\paragraph{Surface–mesh updating per object.}
Let $T_t=(P_t,K)$ denote the camera parameters at time $t$, consisting of the camera pose $P_t$ and intrinsics $K$.
From the depth map $D_t$ and camera parameters $T_t$, we obtain a point cloud
\[
\mathcal{S}_t = \mathrm{BackProject}(D_t,\, T_t).
\]
Each point is assigned to an object via the segmentation of frame $t$, obtained using SAM2~\cite{ravi2025sam} with user-provided point prompts.

All previously stored surface meshes are transformed into the current frame:
\[
\hat{\mathcal{S}}_{t-1}^{(o)} = 
\mathcal{T}_t^{(o)}\!\left( \mathcal{S}_{t-1}^{(o)} \right),
\]
and fused with the newly observed surface points $\mathcal{S}_t^{(o)}$ to produce the updated surface mesh $\mathcal{S}_t^{(o)}$.  
This ensures that surfaces move coherently according to the object’s segmentation and SE(3) motion. Because the system knows which regions are newly observed versus previously visible under the target viewpoint, it can optionally choose to overwrite, refine, or preserve the pre-existing surface regions during fusion.

\begin{figure*}[!t]
\vspace{-0.4cm}
    \centering
    \includegraphics[width=0.98\linewidth]{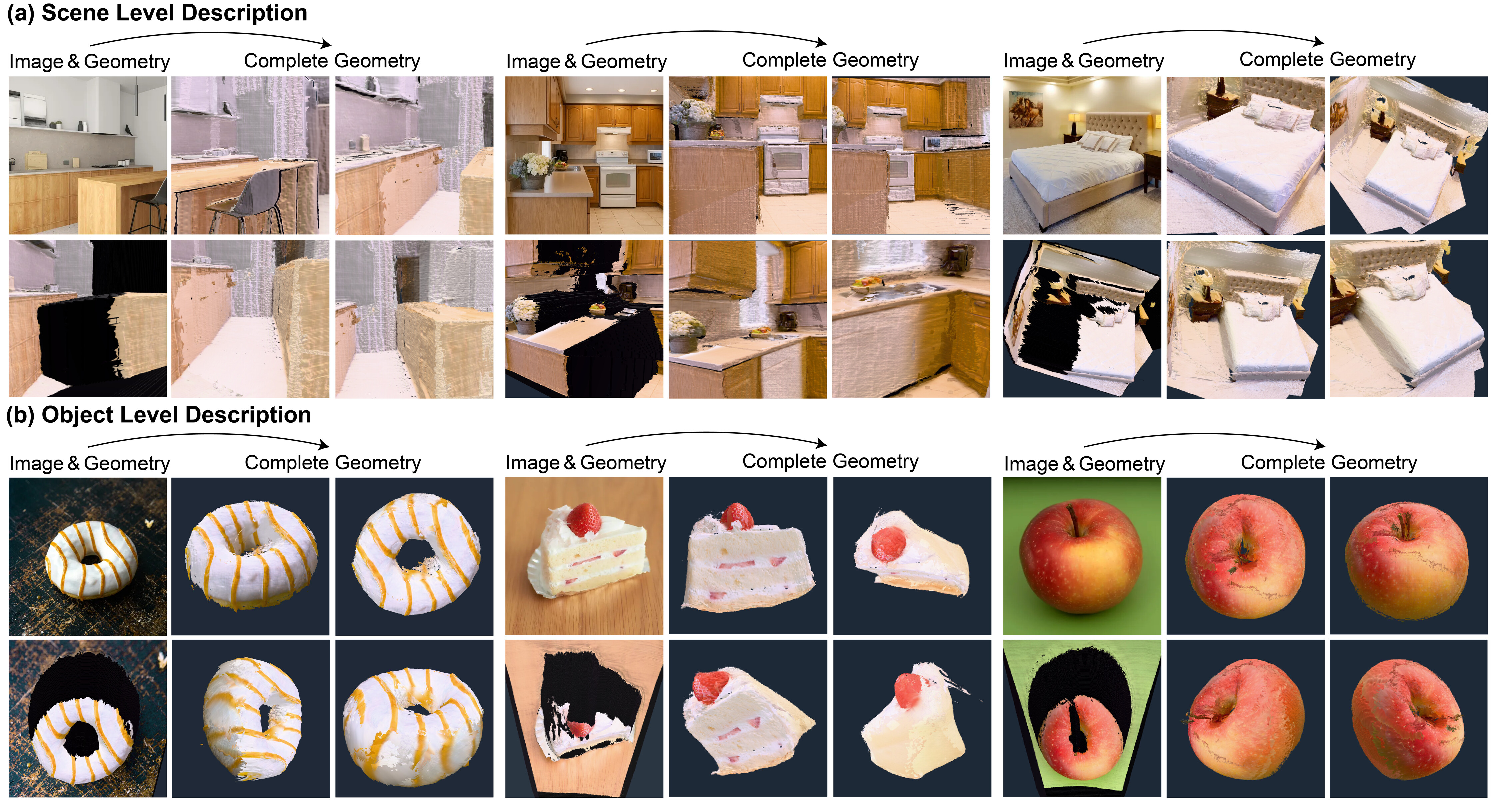}
    \vspace{-0.2cm}
    \captionsetup{labelfont=bf}
    
\caption{
\textbf{Additional Examples of Extracting a Complete Scene Description.}
(a) \emph{Scene-level description:} additional examples of coherent 3D scene layout inferred from single images.
(b) \emph{Object-level description:} further examples of consistent object-surface reconstruction.
}
    \label{fig:more_extracting_geometry}
    \vspace{-0.4cm}
\end{figure*}

% Extracting a Complete Scene Description

\paragraph{Undersurface–mesh updating per object.}
Each object maintains an occupancy volume $\Phi_{t-1}^{(o)}(x) \in \{-1,+1\}$, where $+1$ denotes observed (free) space and $-1$ denotes unobserved space.  
Its zero-level isosurface defines the undersurface mesh $\mathcal{U}_{t-1}^{(o)}$.

Before carving, the occupancy is transported into the current frame:
\[
\hat{\Phi}_{t-1}^{(o)}(x)
= \Phi_{t-1}^{(o)}\!\left((\mathcal{T}_t^{(o)})^{-1}(x)\right).
\]

Given $D_t$, we carve the occupancy volume ray-by-ray using an offset depth $D_t(u) + \varepsilon$ (with $\varepsilon = 3\times10^{-2}$), which guarantees that the undersurface mesh stays just behind the visible surface.
For any voxel $x$ lying in front of this offset depth along ray $u$,
\[
\Phi_t^{(o)}(x) = +1, \qquad
\text{(carved: now observed space)},
\]
and otherwise
\[
\Phi_t^{(o)}(x) = \hat{\Phi}_{t-1}^{(o)}(x).
\]

The updated undersurface mesh is then extracted as the $0$-level isosurface from $\Phi_t^{(o)}$ using marching cubes. This yields a geometry-coherent boundary of the still-unobserved volume behind the object at time $t$.
This occupancy-based update enforces consistency with both past and current observations. Carving with the current depth removes regions inconsistent with the present view, and the monotonicity of $\Phi_t$ prevents reintroducing regions that were already ruled out by earlier viewpoints.

\section{Additional Examples of Extracting a Complete Scene Description}
\label{sec:more_geometry_examples}

We present additional qualitative results demonstrating geometry extraction from single images in Figure~\ref{fig:more_extracting_geometry}. For the first frame, depth $D_0$ is estimated using Depth Anything V2. The next frame depth is sampled as $D_1 \sim \Psi(I_0, D_0, F_{0\!\to\!1}, D_{1}^{\mathrm{sparse}})$, followed by image synthesis $I_1 \sim \Psi(I_0, D_0, F_{0\!\to\!1}, D_1)$. Subsequent steps follow the analogous pattern $D_2 \sim \Psi(I_0, I_1, D_1, F_{1\!\to\!2}, D_{2}^{\mathrm{sparse}})$ and $I_2 \sim \Psi(I_0, I_1, D_1, F_{1\!\to\!2}, D_2)$. The first frame $I_0$ remains fixed, while $I_1$ is continually replaced by the frame closest to the current target.

Throughout inference, the memory module $\mu$ is updated at every step so that geometric information accumulates across frames. This procedure recovers both scene-level geometry, useful for navigation, and object-level geometry, useful for object-centric tasks. For object-level visualization, we show only the surface meshes associated with each object according to the update rules of the memory module $\mu$.

% \newpage
% {
%     \small
%     \bibliographystyle{ieeenat_fullname}
%     \bibliography{main}
% }

% \end{document}

\end{document}